\documentclass{article} 

\usepackage[dvipsnames]{xcolor}
\usepackage{iclr2026_conference}
\usepackage{times}
\usepackage{amsmath,amssymb}
\usepackage{booktabs}
\usepackage{graphicx}
\usepackage{float}
\usepackage{microtype}
\usepackage{enumitem}
\usepackage{hyperref}
\usepackage{url}

\title{Step-Level On-Policy Distillation:\\ Interpolating Between On-Policy Distillation and Supervised Fine-Tuning}

\author{%
\normalsize
\textbf{Changhui Sun$^{1,2}$, Lanbo Liu$^{2}$, Hang Lei$^{2}$, Tong Ling$^{2,3}$}\\
\textbf{Jiahang Xie$^{2,4}$, Zhiyong Zheng$^{2}$, Yujia Wang$^{2}$, Hao Liu$^{2}$}\\
\textbf{Feng Xiao$^{2}$, Lu Liu$^{2}$, Yanlong Du$^{2}$,
Zifeng Cheng$^{1\dagger}$}\\
\textbf{Ziwei Jiang$^{1\dagger}$, Qing Gu$^{1}$}\\
\textnormal{$^{1}$State Key Laboratory for Novel Software Technology, Nanjing University, China}\\
\textnormal{$^{2}$XingYun Lab, HUJING Digital Media \& Entertainment Group}\\
\textnormal{$^{3}$University of Chinese Academy of Sciences, Beijing, China}\\
\textnormal{$^{4}$School of Data Science, Fudan University, Shanghai, China}%
}

\iclrfinalcopy

\begin{document}

\maketitle
\begingroup
\renewcommand{\thefootnote}{\fnsymbol{footnote}}
\footnotetext[2]{Corresponding authors.}
\endgroup

\begin{abstract}
On-policy distillation (OPD) aligns a student model with a teacher's logit distribution on student-generated trajectories. This approach has achieved strong empirical gains and can often surpass conventional off-policy distillation with substantially less data. However, standard token-level OPD can provide only fragmented corrections along an erroneous student trajectory and cannot unfold a complete and correct repair path. Motivated by this limitation, we propose \emph{Step-Level On-Policy Distillation} (SOPD), which combines the long-horizon correction of supervised fine-tuning (SFT) with the on-policy advantage of OPD to provide step-level supervision over complete student-generated trajectories. We show that, at different limits of step length, SOPD reduces to SFT or approximates OPD. Compared with SFT, the teacher responses in SOPD are conditioned on student trajectories and therefore align more closely with student-visited states; compared with OPD, SOPD provides longer-horizon corrections rather than fragmented token-level guidance. Across both reasoning and agent tasks, SOPD substantially outperforms conventional SFT and OPD. For example, on ALFWorld, SOPD improves the average success rate by 13.4 points over Vanilla OPD. We hope this work offers a new perspective for future research on distillation methods.

\end{abstract}

\section{Introduction}
\label{sec:introduction}

Recently, on-policy distillation (OPD)~\citep{agarwal2024policy,qwen3,lu2025onpolicydistillation} has emerged as an effective post-training paradigm for improving the capabilities of large language models (LLMs). Unlike earlier off-policy distillation methods~\citep{taori2023alpaca,guha2025openthoughts}, which train the student on teacher-generated trajectories, OPD allows the student to learn from teacher supervision---namely, predicted logits---on student-generated tokens.

Standard OPD queries a teacher distribution at every token generated by the student. Although this provides relatively dense supervision, it also fragments the teacher's correction signal: in a single OPD rollout, the student receives only one-token guidance at each position rather than a continuous and correct segment-level correction~\citep{trd}. Recent work has revisited the potential of DAgger-style methods for training LLM agents, seeking to preserve SFT's long-horizon supervision while generating more training trajectories from student states~\citep{lauffer2025imitation}. DAgger, however, inherently inserts teacher responses into the student-generated trajectory, substantially changing the trajectory the student would otherwise have produced.

We combine the on-policy property of OPD with the long-horizon supervision of SFT and propose \textbf{Step-Level On-Policy Distillation (SOPD)}. SOPD provides step-level supervision over a complete student-generated trajectory. After the student completes the full trajectory, SOPD asks the teacher to generate one step from the student prefix at every student step. Our method reduces to SFT when the number of steps approaches one and approximates forward-KL OPD when the step length approaches one token.

We show that SOPD outperforms both SFT and OPD across different tasks. For agent tasks, we simply define a step as one actual environment-interaction turn. SOPD substantially improves over both SFT and OPD while reducing the number of interaction rounds. For mathematical reasoning, we define steps using natural reasoning boundaries, and SOPD improves over OPD by an average of 10 percentage points. These results demonstrate the general effectiveness of SOPD for both reasoning and agent tasks.

Moreover, standard OPD requires access to teacher logits, whereas SOPD requires only teacher-generated responses and can therefore be applied more directly to black-box distillation. The total amount of teacher generation per sample is close to the length of one complete response, making the practical cost of black-box calls comparable to standard SFT. In agent tasks, each teacher call generates only one step, so distillation avoids the additional waiting time caused by teacher--environment interaction and tool execution.

Our contributions are:
\begin{itemize}[leftmargin=*,itemsep=2pt,topsep=3pt]
  \item We combine the respective advantages of OPD and SFT and propose Step-Level On-Policy Distillation (SOPD). Our method is also better suited than standard OPD to black-box distillation.
  \item We report results on both agent and reasoning tasks, demonstrating the advantages of SOPD over OPD and SFT.
  \item In two limiting cases, SOPD reduces to standard SFT or approximate forward-KL OPD. This connection between SFT and OPD offers a new perspective on both distillation paradigms.
\end{itemize}

\section{Related Work}
\label{sec:related}

\paragraph{Knowledge distillation.}
Knowledge distillation transfers a teacher distribution into a smaller student~\citep{hinton2015distilling}. Sequence-level distillation trains on teacher-generated sequences and reduces the train--test mismatch created by token-level targets~\citep{kim-rush-2016-sequence}. MiniLLM instead minimizes reverse KL and samples from the student during optimization~\citep{gu2024minillm}. These methods establish sequence generation and student sampling as distinct design choices. SOPD combines student-induced prefixes with teacher-generated local targets.

\paragraph{On-policy language-model distillation.}
Generalized Knowledge Distillation trains the student on its own generations and queries teacher logits at those states~\citep{agarwal2024policy}. Recent work has expanded OPD in several directions. On-policy self-distillation (OPSD) lets the same model serve as student and teacher under ordinary and privileged contexts, achieving stronger reasoning performance and higher training efficiency than off-policy distillation without requiring a separate teacher model~\citep{zhao2026self}. Extrapolative OPD (ExOPD) uses reward extrapolation to relax direct imitation of the teacher, allowing a student that fuses multiple experts to surpass the teachers' performance boundary~\citep{gopd}. SOD, StepOPSD, and SAF-OPD combine step-level distillation signals with GRPO or RLVR to alleviate sparse trajectory rewards and local credit-assignment difficulties~\citep{sod,stepopsd,safopd}. GAD realizes black-box on-policy distillation through a generator--discriminator game and outperforms sequence-level distillation while accessing only teacher-generated text~\citep{ye2025black}. Together, these advances show that OPD has expanded from standard teacher--student logit alignment to self-distillation, surpassing the teacher, integration with RL, and black-box distillation.

At the same time, recent work has identified an inherent limitation of standard OPD. Trajectory-Refined Distillation (TRD) observes that supervision from a single rollout does not contain a complete counterfactual correction trajectory: even with a perfect teacher, token-wise OPD can provide only fragmented corrections along an erroneous student trajectory rather than unfold a complete, correct repair path~\citep{trd}. Later targets remain conditioned on the original erroneous student prefixes instead of prefixes that incorporate earlier teacher corrections. TRD constructs a repair path by rewriting the complete student trajectory; SOPD instead preserves the complete student trajectory while allowing the teacher to generate one coherent local correction within each natural step.

\paragraph{Distillation and guidance for multi-turn agents.}
Action errors compound in interactive agents because every action changes the next observation. ALFWorld instantiates this dependence through textual actions in embodied household tasks~\citep{shridhar2020alfworld}. Existing approaches differ in who generates the training states and whether teacher outputs are executed in the environment.

On-policy Expert Correction (OEC) begins with a student rollout, switches control to an expert partway through the trajectory, and applies SFT to reward-filtered expert suffixes~\citep{lauffer2025imitation}. Agent-RLVR lets the student attempt a task, augments failed attempts with plans, error feedback, or interaction information from an external model, and then applies RL with verifiable environment rewards; the guidance is used only during training~\citep{agentrlvr}. These methods help the learner enter successful regions through expert suffixes or guidance-enhanced reward optimization, rather than producing an independent target at every fixed student state.

TCOD uses a temporal curriculum to control which interval of a multi-turn trajectory is assigned to the student: forward-to-backward expands the student's prefix of control, whereas backward-to-forward begins from a teacher trajectory prefix and moves student control toward the beginning~\citep{tcod}. Guided-OPD chooses a teacher or student action at each environment turn according to a curriculum and anneals the probability of teacher intervention; selected teacher actions are executed, so training visits states induced by a teacher--student mixture~\citep{guidedopd}. ReOPD avoids repeated environment interaction during training by replaying prefixes from pre-collected teacher trajectories, letting the student act at selected steps, and requesting dense teacher supervision~\citep{reopd}.

These methods largely depend on teacher intervention to adjust the generated trajectory, which weakens the on-policy nature of training. SOPD instead provides guidance over a complete student trajectory. At every student-generated step, the teacher produces one independent fragment conditioned on the corresponding student prefix, yielding supervision that remains more faithfully on-policy. Because each teacher call generates only one step, teacher generation does not wait for environment interaction or returned tool results; after the student trajectory is complete, the step-level teacher queries can also be issued in parallel.

\section{Preliminaries: SFT and OPD}
\label{sec:preliminaries}

We first establish notation for the two training paradigms that delimit SOPD. Let $x\sim\mathcal{D}$ be a training prompt, $\pi_\theta$ the student to be optimized, and $\pi_T$ a fixed teacher.

\paragraph{Supervised fine-tuning (SFT).}
SFT applies teacher forcing to a complete teacher or expert sequence $y^T=(y^T_1,\ldots,y^T_L)$:
\begin{equation}
  \mathcal{L}_{\mathrm{SFT}}
  =\mathbb{E}_{(x,y^T)\sim\mathcal{D}_{T}}
  \left[-\frac{1}{L}\sum_{t=1}^{L}
  \log\pi_\theta(y^T_t\mid x,y^T_{<t})\right].
  \label{eq:sft}
\end{equation}
Supervision within a target sequence unfolds along the teacher's own preceding tokens, providing a coherent long-range target. Its contexts, however, come from teacher data rather than the current student, creating an off-policy state-distribution mismatch.

\paragraph{On-policy distillation (OPD).}
OPD first samples $y^S\sim\pi_\theta(\cdot\mid x)$ and then queries the teacher distribution at every student prefix $h_t^S=(x,y^S_{<t})$. A common reverse-KL objective is
\begin{equation}
  \mathcal{L}_{\mathrm{OPD}}
  =\mathbb{E}_{x,\,y^S\sim\pi_\theta}
  \left[\frac{1}{|y^S|}\sum_t
  \mathrm{KL}\!\left(
  \pi_\theta(\cdot\mid h_t^S)\,\|\,
  \pi_T(\cdot\mid h_t^S)\right)\right].
  \label{eq:opd}
\end{equation}
This objective aligns training states with those encountered by the student at inference and supplies dense token-level feedback, but its standard implementation requires teacher logits. Moreover, after the student has entered an erroneous trajectory, the teacher distribution at the next position is still conditioned on an even longer erroneous student prefix. The resulting position-wise targets need not form a correction path that can be executed coherently~\citep{trd}.

\section{Step-Level On-Policy Distillation}
\label{sec:method}

\begin{figure*}[t]
\centering
\includegraphics[width=\textwidth]{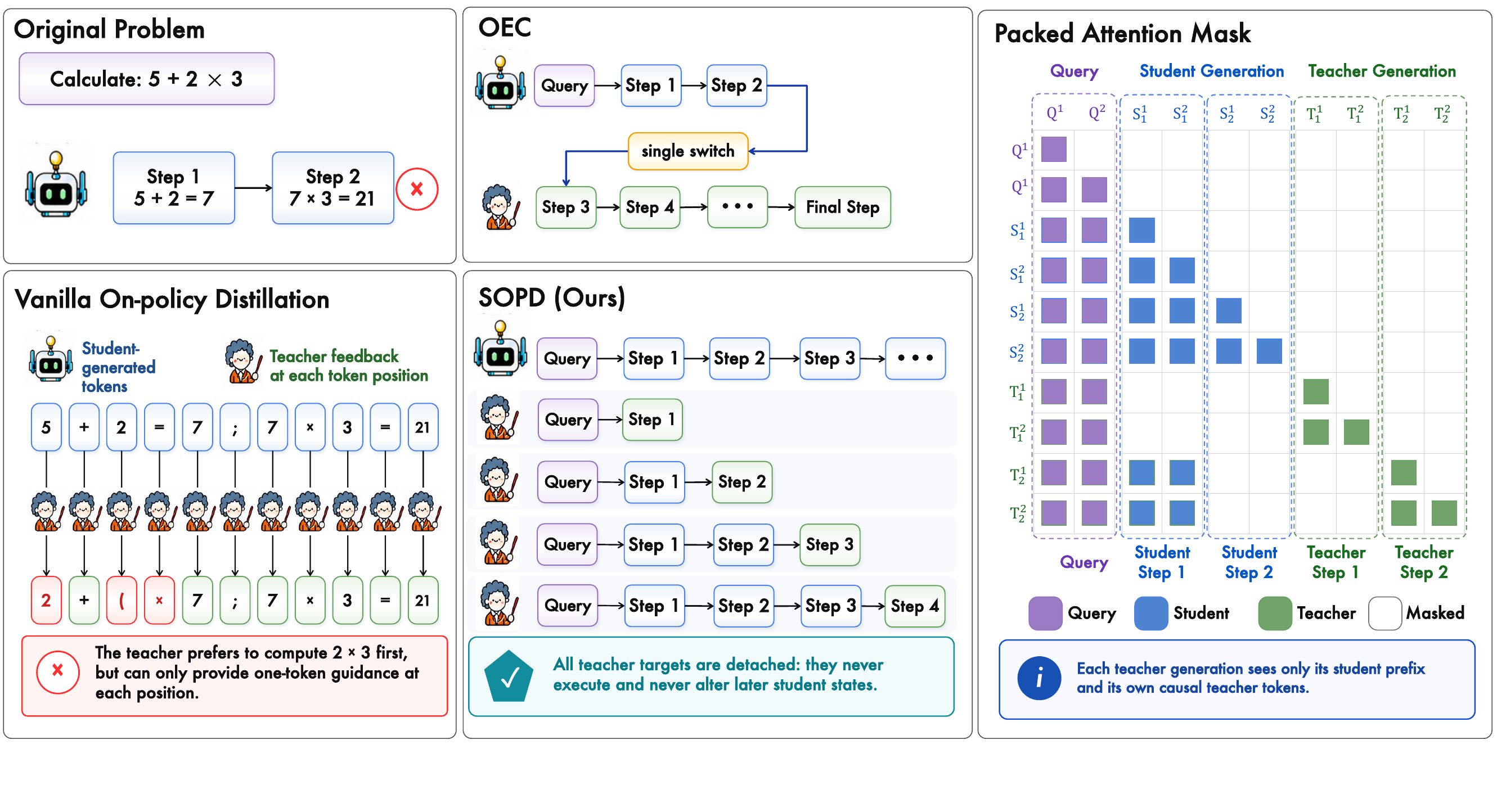}
\caption{\textbf{Overview of Step-Level On-Policy Distillation (SOPD).} The left and center panels compare On-Policy Distillation (OPD), On-Policy Expert Correction (OEC), and SOPD. The right panel illustrates the packed attention mask used for efficient training. We pack the input prompt, the complete student trajectory, and its multiple step-level teacher targets into a single sequence, enabling these step-level supervision instances to be processed in parallel within a single forward pass to ensure training efficiency. Each teacher target can attend only to the input prompt, its corresponding student prefix, and its own preceding causal teacher tokens, while future student steps and other teacher targets remain masked.}
\label{fig:method}
\end{figure*}

SOPD preserves the student state occupancy of OPD while retaining coherent multi-token supervision within each local target. It first partitions a complete student trajectory into natural steps and then asks the teacher to generate one target from the beginning of every step. The student trajectory defines the states between steps; a teacher target unfolds autoregressively only within its own step. Figure~\ref{fig:method} contrasts OEC, token-wise OPD, and SOPD and illustrates the packed attention mask.

\subsection{Student Rollouts Fix the Visited States}

Let $x$ denote a prompt or initial task state. A rollout student $\pi_{\theta_r}$ generates a complete trajectory
\begin{equation}
  \tau_S=(s_1,s_2,\ldots,s_K), \qquad
  s_k \sim \pi_{\theta_r}(\cdot\mid h_k),
  \label{eq:student-trajectory}
\end{equation}
where $s_k$ is one natural step and $h_k$ is the student-visited prefix before that step. For static reasoning, $h_k=(x,s_{<k})$. In an interactive environment, $h_k$ also contains the observations returned after the environment executes the student actions in $s_{<k}$.

The partition follows the structure of each domain. In mathematics, a paragraph boundary or an explicit heading such as ``Step'' or ``Solution'' closes a reasoning step. A 1,024-token safety cap closes a span only when the model emits no natural boundary. In ALFWorld, one assistant response and its parsed action form a step. The environment executes that student action and returns the observation that defines the next prefix.

SOPD records every $h_k$ before teacher generation. It therefore preserves the state distribution induced by the rollout student:
\begin{equation}
  h_k \sim d^{\pi_{\theta_r}}_k.
  \label{eq:student-occupancy}
\end{equation}
Because the rollout has already fixed the complete trajectory, teacher outputs cannot alter $d^{\pi_{\theta_r}}_k$. Thus, every step begins at a state actually visited by the student rather than at one produced by executing a teacher target.

\subsection{Teacher Targets Do Not Alter the Trajectory}

Let $q_T(\cdot\mid h_k)$ denote the teacher distribution over the next natural step from student prefix $h_k$. For every recorded prefix, the teacher generates one target step:
\begin{equation}
  \tilde{s}_k \sim q_T(\cdot\mid h_k).
  \label{eq:teacher-target}
\end{equation}
Each query contains the original prompt and the preceding \emph{student} steps, but no earlier teacher target. This aligns every target with a state the student actually visited and prevents teacher-written text from defining the context of a later target. The teacher supplies generated text only; SOPD requires neither teacher logits nor gradients and therefore supports black-box teachers.

For mathematics, the teacher stops at the same family of natural boundaries used by the partitioner. For agents, all recorded turn states are submitted after the episode. Because no target appears in another target's context, these teacher queries are conditionally independent, can be generated in parallel, and require neither tool execution nor new environment observations.

\subsection{Step Balancing Prevents Long Targets from Dominating}

SOPD trains the student to generate each teacher target from its corresponding student prefix. Let $\mathcal{S}_{B}$ contain every teacher step target collected from rollout batch $B$. We average token cross-entropy within each target and then average over the pooled steps:
\begin{equation}
  \mathcal{L}_{\mathrm{SOPD}}(B)
  = \frac{1}{|\mathcal{S}_{B}|}
    \sum_{(h_k,\tilde{s}_k)\in\mathcal{S}_{B}}
    \frac{1}{|\tilde{s}_k|}\sum_{j=1}^{|\tilde{s}_k|}
    -\log \pi_\theta
    \left(\tilde{s}_{k,j}\mid h_k,\tilde{s}_{k,<j}\right).
  \label{eq:sopd-loss}
\end{equation}
The inner average prevents a long teacher step from dominating shorter decisions, and the outer average assigns equal weight to every collected step. A trajectory with more retained steps consequently contributes more step terms. Under stochastic teacher decoding, one target is a Monte Carlo sample of the length-normalized risk
\begin{equation}
  \mathbb{E}_{\tilde{s}\sim q_T(\cdot\mid h)}
  \left[-\frac{1}{|\tilde{s}|}\log \pi_\theta(\tilde{s}\mid h)\right].
  \label{eq:sampled-step-risk}
\end{equation}
Under greedy decoding, $q_T=\delta_{g_T(h_k)}$, and the step term equals $|g_T(h_k)|^{-1}\mathrm{KL}(\delta_{g_T(h_k)}\|\pi_\theta)$. This length normalization distinguishes Equation~\ref{eq:sopd-loss} from unnormalized sequence-level forward KL.

The packed implementation shares repeated prefixes without changing Equation~\ref{eq:sopd-loss}. It stores the prompt, the student steps needed by every prefix, and all teacher targets in one packed example. Its attention mask lets target $\tilde{s}_k$ attend to $x$, $s_{<k}$, and its own causal target prefix, while blocking $s_{\geq k}$ and every other teacher target. Loss labels cover teacher tokens only.

\subsection{SOPD Connects Two Granularity Limits}

The step granularity locates SOPD between sequence SFT and token-level OPD.

\paragraph{One-step limit: sequence SFT.}
If an entire response is the sole step, then $K=1$, $h_1=x$, and the teacher generates a complete target $\tilde{y}\sim q_T(\cdot\mid x)$. Equation~\ref{eq:sopd-loss} becomes
\begin{equation}
  \mathcal{L}_{K=1}
  =-\frac{1}{|\tilde{y}|}\sum_j
  \log\pi_\theta(\tilde{y}_j\mid x,\tilde{y}_{<j}),
  \label{eq:sequence-limit}
\end{equation}
which is the sequence teacher-forcing loss in Equation~\ref{eq:sft}. Caching the teacher target yields ordinary teacher-data SFT; online generation changes how the data are acquired, not the optimization objective.

\paragraph{One-token limit: a sampled forward-KL OPD objective.}
If every step contains exactly one token, the step at position $t$ begins at student prefix $h_t^S$, and the teacher samples $\tilde y_t\sim q_T(\cdot\mid h_t^S)$. Taking expectation over this sample gives
\begin{align}
  \mathbb{E}_{\tilde y_t\sim q_T}
  [-\log\pi_\theta(\tilde y_t\mid h_t^S)]
  &= H\!\left(q_T(\cdot\mid h_t^S),\pi_\theta(\cdot\mid h_t^S)\right)\notag\\
  &= H\!\left(q_T(\cdot\mid h_t^S)\right)
  +\mathrm{KL}\!\left(q_T(\cdot\mid h_t^S)\|\pi_\theta(\cdot\mid h_t^S)\right).
  \label{eq:token-limit}
\end{align}
The teacher entropy is constant with respect to $\theta$, so minimizing expected cross-entropy is equivalent to minimizing forward KL on student-visited prefixes. A single sampled teacher token is a Monte Carlo estimate of that objective. A greedy teacher instead yields hard imitation of the teacher's argmax action, not an unbiased estimate of the full-distribution forward KL.

A natural step lies between these limits. Its start remains student-selected, retaining OPD's student-state coverage; its multiple target tokens unfold along the teacher's own prefix, retaining the locally coherent supervision of SFT. The teacher can therefore express an executable correction within a step without changing the state at which the next student step was observed.

\subsection{Relation to Other Distillation Procedures}

Off-policy sequence distillation trains on teacher-generated trajectories~\citep{kim-rush-2016-sequence}. Standard OPD queries teacher probabilities along student outputs and minimizes a token-level divergence~\citep{agarwal2024policy}. SOPD retains student-induced states but replaces logit access with local teacher generation.

ExOPD uses reward extrapolation to enable a student to exceed a single teacher~\citep{gopd}; SOPD uses no reward in its objective. TCOD assigns control intervals in a multi-turn trajectory through a temporal curriculum~\citep{tcod}; Guided-OPD allows teacher actions to enter the environment according to a curriculum~\citep{guidedopd}; and ReOPD replays offline teacher prefixes instead of interacting with the environment throughout training~\citep{reopd}. SOPD instead keeps the natural temporal order, queries every student-visited step, and never executes a teacher target. Step granularity together with student-state anchoring defines the method.

\section{Experiments}
\label{sec:experiments}
\label{sec:evaluation}

We compare SOPD and OPD in two settings with long student-induced trajectories. ALFWorld partitions an interactive trajectory into environment turns, whereas mathematical reasoning partitions one response into reasoning steps. Both evaluations measure end-to-end outcomes and use the corresponding natural step structure during training.

\subsection{ALFWorld Agent Interaction}

\paragraph{Setup.}
Following TCOD~\citep{tcod}, we use a domain-RL-trained Qwen2.5-7B teacher to distill a Qwen2.5-3B-Instruct student. The Explorer samples batches of 16 tasks, each rollout contains at most 30 environment turns, and the Trainer consumes 64 experiences per batch with maximum staleness two. We evaluate on all 140 Valid Seen, 134 Valid Unseen, and 121 Hard tasks at temperature 0.4 with a 4,096-token response cap. Success rate measures task completion, and Rounds counts environment interactions.

\begin{table*}[t]
\centering
\small
\setlength{\tabcolsep}{4.6pt}
\caption{\textbf{ALFWorld results with a 3B student and 7B RL teacher.} SR denotes success rate (\%). TCOD supplies all rows except SOPD~\citep{tcod}. Arrows show the absolute change from Vanilla OPD; bold marks the best student result in each column.}
\label{tab:alfworld-main}
\resizebox{\textwidth}{!}{%
\begin{tabular}{lcccccc}
\toprule
& \multicolumn{2}{c}{Valid Seen} & \multicolumn{2}{c}{Valid Unseen} & \multicolumn{2}{c}{Hard} \\
\cmidrule(lr){2-3}\cmidrule(lr){4-5}\cmidrule(lr){6-7}
Method & SR $\uparrow$ & Rounds $\downarrow$ & SR $\uparrow$ & Rounds $\downarrow$ & SR $\uparrow$ & Rounds $\downarrow$ \\
\midrule
Qwen2.5-7B-RL (Teacher) & 85.71 & 10.61 & 76.87 & 13.06 & 6.61 & 27.31 \\
\midrule
Qwen2.5-3B Zero-Shot & 7.86 & 28.73 & 2.24 & 29.63 & 0.83 & 29.88 \\
Supervised fine-tuning (SFT) & 32.14 & 22.85 & 25.37 & 24.16 & 4.96 & 29.12 \\
Vanilla OPD & 65.72 & 14.73 & 60.45 & 16.21 & 10.74 & 28.64 \\
TCOD (B2F) & 77.86 $\uparrow$12.14 & 12.57 $\downarrow$2.16 & 70.90 $\uparrow$10.45 & 14.56 $\downarrow$1.65 & \textbf{13.22} $\uparrow$2.48 & 28.16 $\downarrow$0.48 \\
TCOD (F2B) & 81.43 $\uparrow$15.71 & 11.76 $\downarrow$2.97 & 79.19 $\uparrow$18.74 & 12.47 $\downarrow$3.74 & 9.92 $\downarrow$0.82 & 28.57 $\downarrow$0.07 \\
\textbf{SOPD} & \textbf{84.29} $\uparrow$18.57 & \textbf{11.20} $\downarrow$3.53 & \textbf{82.09} $\uparrow$21.64 & \textbf{11.88} $\downarrow$4.33 & 10.74 $\leftrightarrow$0.00 & \textbf{28.15} $\downarrow$0.49 \\
\bottomrule
\end{tabular}
}
\end{table*}

\paragraph{The SOPD run leads on Seen and Unseen interaction.}
Relative to Vanilla OPD, SOPD raises Seen success from 65.72 to 84.29 and Unseen success from 60.45 to 82.09 while reducing mean rounds by 3.53 and 4.33, respectively. Among the student methods, SOPD records both the highest success and the fewest rounds on these two splits. Its Unseen success exceeds TCOD-F2B by 2.90 points.

\paragraph{Performance across difficulty splits.}
On Hard, SOPD reaches 10.74\% success while recording 28.15 mean rounds, the shortest interaction trajectory among all student methods. Together with the substantial gains on Seen and Unseen, these results show that SOPD learns common and unseen interaction states more effectively while maintaining stable behavior on difficult tasks.

\begin{figure*}[t]
\centering
\includegraphics[width=.98\textwidth]{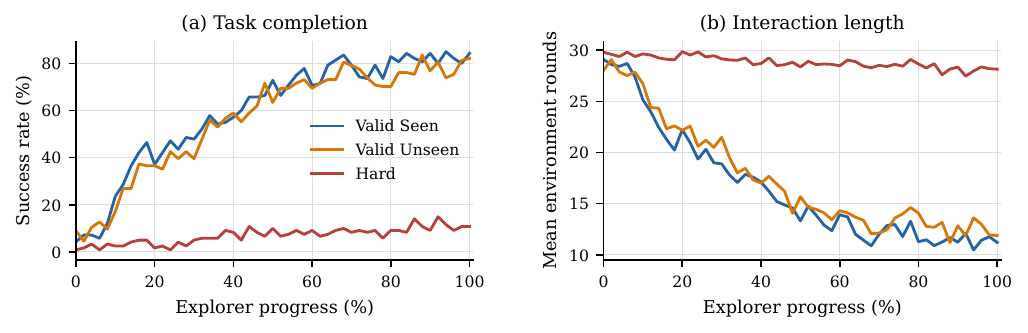}
\caption{\textbf{ALFWorld full-split evaluation over normalized Explorer progress.} Seen and Unseen success improve over training while interaction length decreases overall. The curves summarize full-split evaluations at the saved checkpoints.}
\label{fig:alfworld-dynamics}
\end{figure*}

Figure~\ref{fig:alfworld-dynamics} shows the full-split training trajectory corresponding to Table~\ref{tab:alfworld-main}. Seen and Unseen success improve overall during training as their interaction lengths decrease, while the Hard split remains comparatively stable. This trend agrees with the main table and shows that SOPD improves task completion while learning more efficient interaction policies.

\begin{figure*}[t]
\centering
\includegraphics[width=.99\textwidth]{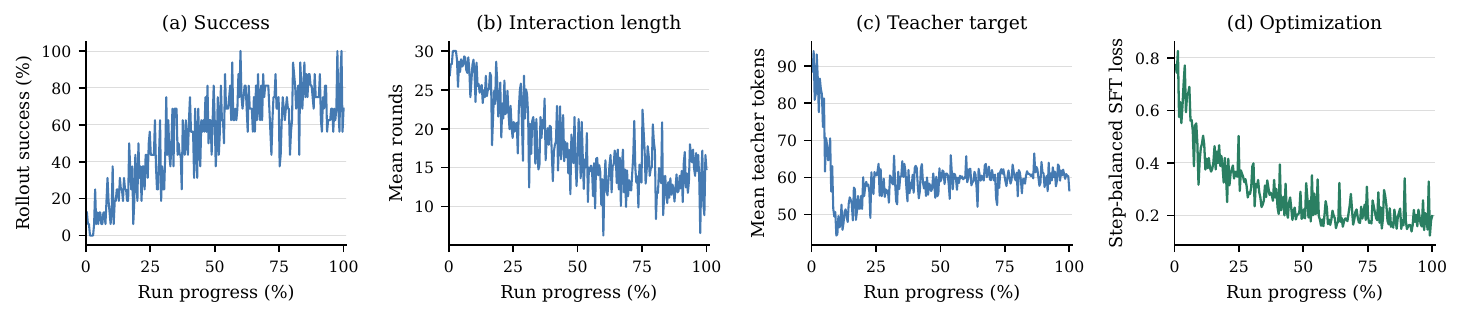}
\caption{\textbf{ALFWorld training diagnostics following TCOD's multi-panel analysis.} (a) Explorer-batch task success; (b) mean environment rounds; (c) mean tokens per generated teacher turn target; and (d) step-balanced SFT loss in the Trainer. The horizontal axis normalizes each signal to the endpoint of its corresponding component.}
\label{fig:alfworld-training}
\end{figure*}

Following TCOD's multi-panel analysis~\citep{tcod}, Figure~\ref{fig:alfworld-training} reports SOPD-specific online signals. The generation-only objective computes no KL quantity. Explorer curves report the current 16-task rollout batches, whereas the optimization panel follows the Trainer. These online signals complement the full-split evaluations in Figure~\ref{fig:alfworld-dynamics}.

\subsection{Mathematical Reasoning}

\paragraph{Setup.}
We primarily use the Qwen3-4B non-thinking model~\citep{qwen3}. The student is initialized from Qwen3-4B non-thinking, and the domain teacher is obtained by applying reinforcement learning to the same base model on domain-specific data. Following G-OPD~\citep{gopd}, we filter DeepMath~\citep{deepmath} to retain 57,000 examples with difficulty level at least 6 and use this set as the mathematical RL data. We then apply Group Relative Policy Optimization (GRPO)~\citep{deepseekmath} to obtain Qwen3-4B-Non-Thinking-RL-Math. The teacher receives reward 1.0 when the final answer is correct and 0.0 otherwise. Appendix~\ref{app:training} reports the complete teacher-training configuration.

For distillation, SOPD partitions responses at natural reasoning boundaries, whereas standard OPD optimizes sampled-token reverse KL. Both methods use the same filtered DeepMath prompt pool and sample one student response for each prompt. Detailed SOPD and OPD hyperparameters, including the step-partition rule, are given in Appendix~\ref{app:training}.

We evaluate mathematical reasoning on four competition-level benchmarks: AIME24~\citep{aime2024}, AIME25~\citep{aime2025}, HMMT25-Feb, and HMMT25-Nov~\citep{hmmt25}. For every evaluation, we set temperature and top-$p$ to 1.0 and the maximum generation length to 32,768 tokens. We sample 32 solutions for each problem and report mean accuracy on each benchmark. Answer correctness is determined by the rule-based \texttt{Math-Verify}\footnote{\url{https://github.com/huggingface/Math-Verify}} verifier.

\begin{table*}[t]
\centering
\small
\setlength{\tabcolsep}{6.5pt}
\caption{\textbf{Comparison of OPD and SOPD using the same-size teacher--student pair.} Teacher denotes Qwen3-4B-Non-Thinking-RL-Math, and Student denotes the initial Qwen3-4B non-thinking model. The first three rows are reference results reported by G-OPD~\citep{gopd}. Values are mean accuracy (\%); $\mathrm{Avg}_4$ averages the four benchmarks, and bold marks the best student method in each column.}
\label{tab:math-main}
\resizebox{\textwidth}{!}{%
\begin{tabular}{lccccc}
\toprule
Method & AIME24 & AIME25 & HMMT25 (Feb.) & HMMT25 (Nov.) & $\mathrm{Avg}_4$ \\
\midrule
Teacher & 58.0 & 54.6 & 32.5 & 38.9 & 46.0 \\
$+$ continued RL (100 updates) & 60.9 & 55.6 & 32.8 & 38.4 & 46.9 \\
Student & 21.5 & 21.9 & 10.0 & 8.0 & 15.4 \\
\midrule
OPD & 61.9 & 57.0 & 32.5 & 39.6 & 47.7 \\
\textbf{SOPD} & \textbf{71.8} & \textbf{67.4} & \textbf{38.9} & \textbf{52.7} & \textbf{57.7} \\
\bottomrule
\end{tabular}
}
\end{table*}

\paragraph{The SOPD run leads on all four benchmarks.}
SOPD raises mean accuracy over OPD by 9.9 points on AIME24, 10.4 points on AIME25, 6.4 points on HMMT25-Feb, and 13.1 points on HMMT25-Nov. Its four-benchmark average reaches 57.7, compared with 47.7 for OPD. Each cell aggregates 32 decoding samples for each of 30 problems, providing a comprehensive evaluation of the final checkpoint.

\begin{figure*}[t]
\centering
\includegraphics[width=.96\textwidth]{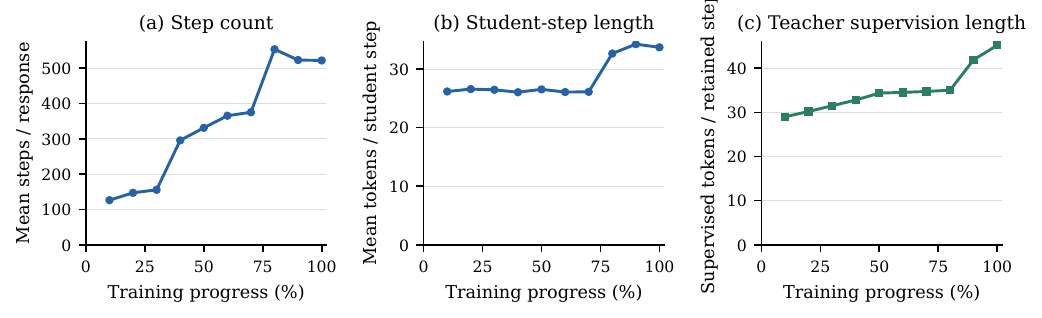}
\caption{\textbf{SOPD step structure over mathematical training.} (a) Mean natural steps per AIME24 evaluation response; (b) mean student tokens per natural step; and (c) mean supervised teacher tokens per retained step during training. All boundaries use the semantic-step rule and 1,024-token safety cap.}
\label{fig:step-stats}
\end{figure*}

\paragraph{The learned response structure changes during training.}
Figure~\ref{fig:step-stats} tracks the quantities that define SOPD's supervision units. As training progresses, the number of natural reasoning steps generated by the student increases substantially, while the length of each step remains broadly stable. This indicates that the model extends its solution process mainly by organizing more structured reasoning steps. Mean teacher supervision grows from 29.0 tokens at update 1 to 35.0 at update 8, then increases to 41.9 and 45.2 at the final two updates. Together, the three statistics show that later checkpoints both organize more reasoning steps and receive richer teacher targets at each local state.

\section{Conclusion}

SOPD turns student-visited intermediate states into black-box teacher queries. It preserves the complete student trajectory, generates one teacher target at every natural step, and optimizes a step-balanced cross-entropy loss. At response-level granularity, the method recovers sequence SFT; at token-level granularity, it corresponds to a sampled forward-KL OPD objective. Results in mathematical reasoning and ALFWorld show that SOPD provides coherent, dense local supervision without allowing the teacher to control later states, yielding consistent gains across four mathematical benchmarks and the ALFWorld Seen and Unseen splits.

\section*{Reproducibility Statement}

Section~\ref{sec:experiments} specifies the model pairs, training data, rollout protocol, optimization settings, and evaluation suites used in the main experiments. Appendix~\ref{app:training} gives the per-update batch sizes, learning rates, decoding parameters, response limits, evaluation sample counts, seeds, and ALFWorld interaction limits. The submission source includes the figure-generation scripts and the aggregated records used to produce the reported tables and figures.

\bibliography{references}
\bibliographystyle{iclr2026_conference}

\appendix

\section{Training and Evaluation Details}
\label{app:training}

\subsection{Mathematical Reasoning}

The mathematical runs use Qwen3-4B non-thinking as the base student and Qwen3-4B-Non-Thinking-RL-Math-Step500 as the teacher. We construct the teacher following the mathematical GRPO configuration of G-OPD~\citep{gopd}. Specifically, we retain 57,000 DeepMath examples with difficulty level at least 6, assign reward 1.0 when the final answer is correct and 0.0 otherwise, and train for 500 optimizer updates. Table~\ref{tab:teacher-grpo} lists the complete teacher-training configuration used in our experiments.

\begin{table}[H]
\centering
\small
\setlength{\tabcolsep}{7pt}
\caption{GRPO hyperparameters used to train the mathematical domain teacher, following G-OPD~\citep{gopd}.}
\label{tab:teacher-grpo}
\begin{tabular}{lc}
\toprule
Hyperparameter & Value \\
\midrule
Training batch size & 128 \\
Micro-batch size & 128 \\
Rollouts per prompt & 8 \\
Maximum prompt length & 2,048 \\
Maximum response length & 16,384 \\
Temperature & 1.0 \\
Top-$p$ & 1.0 \\
Learning rate & $1\times10^{-6}$ \\
Optimizer updates & 500 \\
KL coefficient & 0.0 \\
\bottomrule
\end{tabular}
\end{table}

SOPD and OPD then consume the same fixed order of prompts from the filtered DeepMath pool. Each optimizer update uses 1,024 prompts and one student rollout per prompt. Both methods run for ten updates, so each processes 10,240 prompt instances and 10,240 student rollouts. The maximum prompt and response lengths are 2,048 and 32,768 tokens, respectively; student rollouts use temperature 1.0 and top-$p$ 1.0. SOPD uses learning rate $2\times10^{-6}$ and generates each teacher step greedily at temperature 0. Standard OPD uses learning rate $10^{-5}$ and sampled-token reverse KL.

For SOPD, a blank-line paragraph boundary or an explicit reasoning heading such as ``Step'' or ``Solution'' terminates a natural step. If no such boundary appears, a 1,024-token safety cap closes the span. Every retained student prefix receives one independently generated teacher step, and the loss is averaged first over tokens within each target and then over retained steps.

The terminal evaluation sends 32 independent generations for each of 30 problems in AIME24, AIME25, HMMT25-Feb, and HMMT25-Nov. Each method therefore contributes 960 responses per benchmark. Evaluation uses temperature 1.0, top-$p$ 1.0, and a 32,768-token response cap. Every response is a separate request with seed $42+32i+j$ for problem $i$ and sample $j$. The evaluator extracts the final boxed answer and applies rule-based symbolic verification with \texttt{Math-Verify}. The result extractor checks all numerators and denominators against the durable summaries.

\subsection{ALFWorld}

We use the Qwen2.5-3B-Instruct student and `langfeng01/GiGPO-Qwen2.5-7B-Instruct-ALFWorld' teacher from the TCOD model pairing. The Explorer samples 16 tasks with one rollout each. The Trainer consumes 64 experiences per batch with learning rate $10^{-6}$, temperature 1.0, staleness two, a 512-token training response cap, a 10,240-token prompt cap, and at most 30 environment turns. The released asynchronous launcher lets Explorer and Trainer counters advance independently.

Full evaluation covers Valid Seen-140, Valid Unseen-134, and Hard-121. It samples at temperature 0.4 with a 4,096-token response cap. We report the task-level success mean and mean environment rounds.

\section{LLM Usage}

Generative AI tools were used to assist language editing, translation, figure-layout iteration, and LaTeX preparation. The authors determined the research claims, methods, experiments, and conclusions, and manually verified the final manuscript and reported results.

\end{document}